\documentclass[aps,pre,groupedaddress,showpacs]{revtex4-1}
\usepackage{amsmath,amssymb,subfigure,fancybox,txfonts,bm,color,wrapfig}
\usepackage{siunitx,numprint}
\usepackage[dvipdfmx]{graphicx}

\begin{document}
\newcommand{\vc}{\mathbf}
\newcommand{\gvc}[1]{\mbox{\boldmath $#1$}}
\newcommand{\emf}[1]{{\gtfamily \bfseries #1}}
\newcommand{\fracd}[2]{\frac{\displaystyle #1}{\displaystyle #2}}
\newcommand{\ave}[1]{\left< #1 \right>}
\newcommand{\red}[1]{\textcolor{red}{#1}}
\newcommand{\blue}[1]{\textcolor{blue}{#1}}
\newcommand{\green}[1]{\textcolor[rgb]{0,0.6,0}{#1}}

\newcommand{\del}[2]{\frac{\partial #1}{\partial #2}}
\newcommand{\dev}[2]{\frac{\text{d} #1}{\text{d}#2}}
\newcommand{\mdev}[3]{\frac{\text{d}^{#3} #1}{\text{d}#2^{#3}}}
\newcommand{\pdev}[2]{{\text{d} #1}/{\text{d}#2}}
\newcommand{\intd}[1]{\text{d} {#1}}

\newcommand{\eng}[1]{#1}
 \newcommand{\jpn}[1]{}
\newcommand{\subti}[1]{\begin{itemize} \item {\bf #1} \end{itemize}}

\newcommand{\Real}{\operatorname{Re}}
\newcommand{\Imag}{\operatorname{Im}}

\newcommand{\am}{{\bm a}}
\newcommand{\bb}{{\bm b}}
\newcommand{\ff}{{\bm f}}
\newcommand{\pp}{{\bm p}}
\newcommand{\rr}{{\bm r}}
\newcommand{\sm}{{\bm s}}
\newcommand{\tm}{{\bm t}}
\newcommand{\uu}{{\bm u}}
\newcommand{\ww}{{\bm w}}
\newcommand{\xx}{{\bm x}}
\newcommand{\yy}{{\bm y}}
\newcommand{\zz}{{\bm z}}
\newcommand{\Model}{{\mathbb M}}
\newcommand{\RR}{{\mathbb R}}
\newcommand{\NN}{{\mathbb N}}
\newcommand{\oomega}{\mbox{\boldmath $\omega$}}
\newcommand{\WW}{{\bm W}}
\newcommand{\EE}{\mbox{\boldmath $E$}}
\newcommand{\FF}{\mbox{\boldmath $F$}}
\newcommand{\KK}{\mbox{\boldmath $K$}}
\newcommand{\GG}{\mbox{\boldmath $G$}}
\newcommand{\CC}{\mbox{$\hat{C}$}}
\newcommand{\II}{\mbox{$\hat{I}$}}
\newcommand{\HH}{\mbox{$\hat{H}$}}
\newcommand{\MM}{\mbox{$\hat{M}$}}
\newcommand{\Am}{\mbox{$\hat{A}$}}
\newcommand{\PP}{\mbox{$\hat{P}$}}
\newcommand{\QQ}{\mbox{$\hat{Q}$}}

\title { 
Optical skin: Sensor-integration-free multimodal flexible sensing
}

\author{Sho Shimadera$^1$}
\author{Kei Kitagawa$^2$}
\author{Koyo Sagehashi$^1$}
\author{Tomoaki Niiyama$^3$}
\author{Satoshi Sunada$^{3,4}$}
\email{sunada@se.kanazawa-u.ac.jp}

\affiliation{
$^1$Graduate School of Natural Science and Technology, Kanazawa University, 
Kakuma-machi, Kanazawa, Ishikawa 920-1192, Japan \\
$^2$College of Science and Engineering, Kanazawa University, 
Kakuma-machi, Kanazawa, Ishikawa 920-1192, Japan \\
$^3$Faculty of Mechanical Engineering, Institute of Science and
Engineering, Kanazawa University\\
Kakuma-machi, Kanazawa, Ishikawa 920-1192, Japan \\
$^4$Japan Science and Technology Agency (JST), PRESTO, 4-1-8 Honcho,
 Kawaguchi, Saitama 332-0012, Japan\\
}
\date{\today}

\begin{abstract}
The biological skin enables animals to sense various stimuli. 
Extensive efforts have been made recently to develop smart skin-like sensors to extend the capabilities of biological skins; however, simultaneous sensing of several types of stimuli in a large area remains challenging because this requires large-scale sensor integration with numerous wire connections.
We propose a simple, highly sensitive, and multimodal sensing approach, which does not require integrating multiple sensors. 
The proposed approach is based on an optical interference technique, which can encode the information of various stimuli as a spatial pattern. 
In contrast to the existing approach, the proposed approach, combined with a deep neural network, enables us to freely select the sensing mode according to our purpose. 
%
As a key example, we demonstrate {\it simultaneous} sensing mode of three different physical quantities---contact force, contact location, and temperature---using a {\it single} soft material without requiring complex integration. 
%
Another unique property of the proposed approach is spatially continuous sensing with ultrahigh resolution of few tens of micrometers, which enables identifying the shape of the object in contact.
Furthermore, we present a haptic soft device for a human--machine interface.
The proposed approach encourages the development of high-performance optical skins.
\end{abstract}

\maketitle

\section*{Introduction}
Skin is the largest organ in the human body and enables humans to sense 
various physical stimuli to obtain information regarding their surrounding environment. 
Advanced technologies to build electrical or optical components on soft and stretchable substrates, i.e., flexible electronics \cite {Lipomi:2011aa,Parke1500661,Pue1700015,Chen:2017aa,Someya:2016aa,Zhu:2019aa,Gong2014,C8NH00125A} and flexible photonics \cite{Chunhuan:aa,Choi:2016aa,Larson1071,Ramuz2012_AdvMat,D0NH00393J,Tomoyuki:aa,doi:10.1063/1.3386588,Zhaoeaai7529,Xueaaw6304,8412516,LeiZhang2020Opto-ElectronicAdvances,Xie:2019aa}, have been developed to mimic or extend the capabilities of biological skin. These technologies have led to new opportunities for numerous technical applications, including wearable electronics \cite{Gong2014,POLYGERINOS2015135,Ray:2019aa,Bariya:2018aa}, augmented reality \cite{Yu:2019aa}, prosthetic skins \cite{Chortos:2016aa}, and soft robotics \cite{Huang:2019aa}. 
In the past decades, significant progress has been made to develop sensing capabilities with high sensitivity, high resolution, and fast response in flexible and stretchable substrates with a large area \cite{Li:2020aa}. 
Different sensing mechanisms have been proposed and demonstrated, including electrical signal transduction strategies such as resistive \cite{C8NH00125A}, capacitance \cite{Lipomi:2011aa}, piezoelectric\cite{Parke1500661}, and triboelectric\cite{Pue1700015} methods, 
or a detection strategy based on optical waveguide structures, such as 
stretchable waveguides \cite{Zhaoeaai7529}, optical fibers \cite{Xueaaw6304,8412516,LeiZhang2020Opto-ElectronicAdvances}, and fiber Bragg gratings \cite{Xu:18,s20051312}.
Although these sensing mechanisms can play a crucial role in detecting different types of physical stimuli, 
most currently available sensors focus on 
detecting only a single physical stimulus
and cannot distinguish between multiple stimuli simultaneously. 
Multimodal sensing capability, i.e., simultaneous sensing of different types of stimuli applied to sensors, is important for robots to efficiently perceive the physical world. 
Recently, considerable effort has been devoted to develop multimodal sensing or multi-functional sensing \cite{Li:2020aa,Parke1500661,Xie:2019aa,Din:2017aa,Truby:2018aa,Thurutheleaav1488,Kimeabc6878,7803400,Wang:2020aa}.
Such cases generally require multiple sensing elements with different sensing mechanisms to be integrated in a single sensing platform, which may need complex fabrication processes and/or may experience interference from other stimuli.
Furthermore, to attain high-level perception such as object recognition, the spatial distribution of physical stimuli over a large-scale area should be measured with high spatial resolution. 
A previous approach to achieve high spatial sensing capability was based on integration of a large number of sensors to form a sensor matrix with numerous wire connections; however, this usually requires high integration complexity.
Although a vision-based tactile sensing approach using a marker displacement is easy to manufacture \cite{s21051920}, the spatial resolution is limited, and it is difficult to simultaneously detect other parameters such as temperature.  

Here, we propose a sensor-integration-free, flexible, high-sensitive, and multimodal sensing approach, which does not require users to package or integrate different sensing materials; instead, 
the users can utilize a soft material  as a multimodal sensing element 
without the formation of sensor arrays or sensor matrices and with minimal wiring. 
Consequently, it is potentially capable of large-scale spatial multimodal sensing.   
The unique feature of the proposed approach is based on the use of the optical interference phenomenon caused in a soft material, which is highly sensitive to external stimuli 
and allows the encoding of different stimuli as a spatial interference pattern. 
By using a data-driven decoding technique, users can freely decode various stimulus signals {\it simultaneously}. 
%
%
As a demonstration, we show the simultaneous sensing mode of different physical quantities and recognition mode for the shape of the contacting object.
We also present a haptic soft-interface device. 
Our sensing approach reveals a novel pathway for high-performance skin-like sensors. 
%
 
\section*{Results}
\begin{figure}[htbp]
\centering\includegraphics[width=16cm]{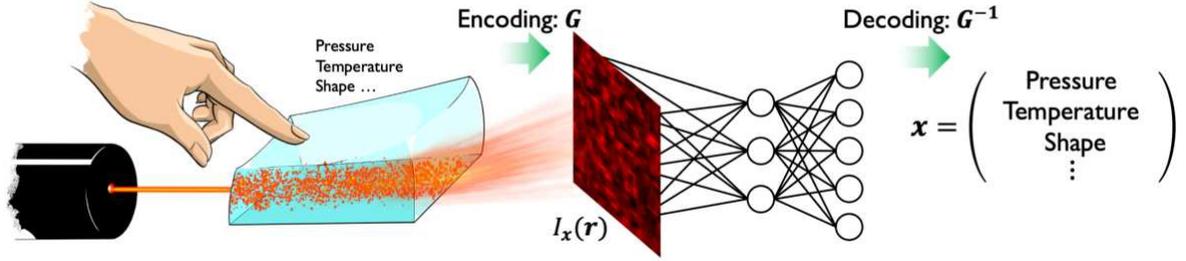}
\caption {\label{fig_concept}
{\bf Conceptual schematic of the proposed soft sensing approach.}
The optical scattering phenomenon inside a soft material 
can induce a complex interference pattern, i.e., speckle pattern, which is 
highly sensitive to external stimuli to the material. 
The information of the external stimuli, $\xx$, can be encoded as the speckle pattern, 
$I_{\xx}(\rr)$. 
The proposed sensing approach is based on the speckle encoding and decoding 
using machine learning. 
}
\end{figure}

{\bf Optical multimodal sensing methodology.}
The proposed sensing approach is based on optical scattering from soft materials
when the material is irradiated with laser light [Fig.~\ref{fig_concept}]. 
The optical scattering produces a complex interference pattern, referred to as 
a speckle pattern \cite{goodman_book_statistical_optics,Goodman:76}, 
on an observation plane. 
The speckle pattern is highly sensitive to the scattering process in a material; 
thus, it can contain various information regarding the deformation of a soft material.  
Although speckle-based techniques have been used in sensing a single physical signal \cite{Murray:19,Fujiwara:17}, deformation field measurement \cite{Fricke-Begemann:2003aa}, three-dimensional shape measurement \cite{Schaffer:10}, and in spectrometers \cite{Redding:13}, this study presents the first demonstration of multimodal sensing in soft materials.

Here, we introduce a vector representing the physical parameters of the external stimuli to the soft material as $\xx = (x_1, x_2, \cdots, x_M) \in \RR^M$.
The speckle intensity pattern measured at the position $\rr$ 
on an observation plane is denoted as $I_{\xx}(\rr)$.
Note that the stimulus information is optically encoded in a high-dimensional feature space as a spatial pattern, $\GG: \xx \rightarrow I_{\xx}(\rr)$. 
The spatial pattern can be regarded as the {\it optical neural response} to an external stimuli.
We can decode $\xx$ from the neural response pattern $I_{\xx}(\rr)$ by identifying an inverse function, represented by $\GG^{-1}: I_\xx(\rr) \rightarrow \xx$.
This can be achieved using a learning-based model.
Therefore, our sensing approach is a model-free (data-driven) approach, which does not require detailed theoretical models of the soft material and optical scattering.
In addition, note that this approach does not require the integration of different types of sensors to detect multiple parameters, $\xx$; a single soft material acts as a sensing unit to separately estimate the multimodal stimulus information simultaneously.

\begin{figure}[htbp]
\centering\includegraphics[width=17cm]{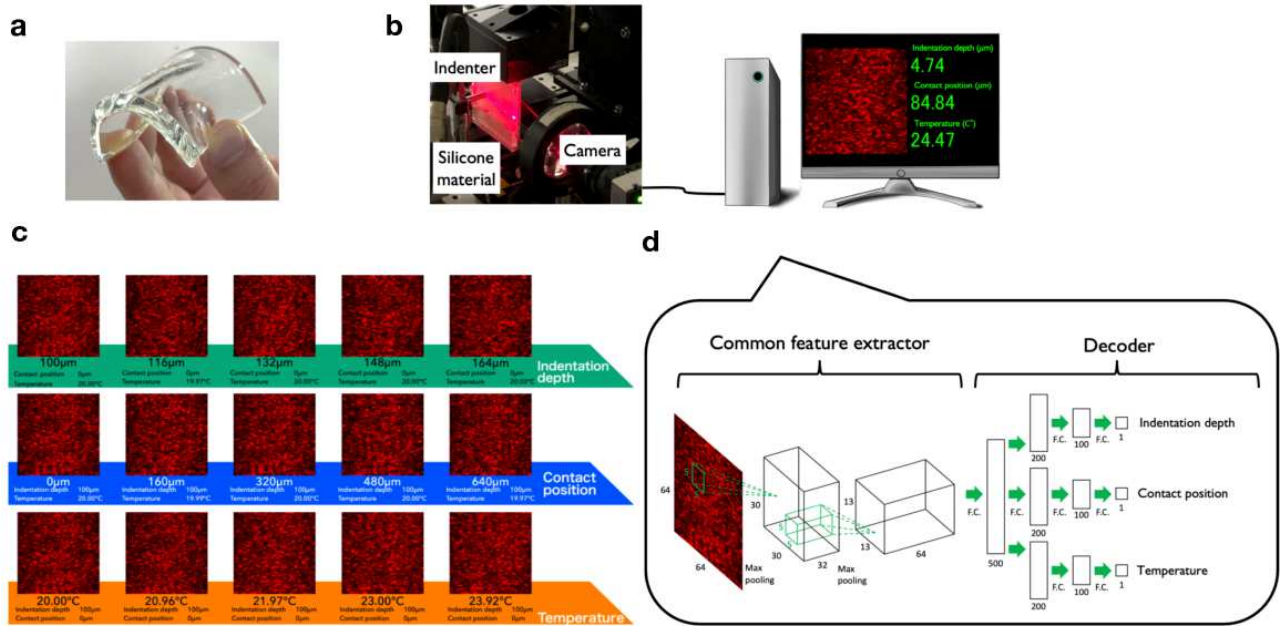}
\caption{\label{fig_exset}
{\bf Experiment of speckle encoding and decoding.} {\bf (a)} Silicone material used in the experiment. 
{\bf (b)}  Experimental setup.
The speckle pattern is measured with a digital camera and processed with the 
network architecture shown in (d).  
The decoded information $\xx$ (indentation depth, contact position, and temperature in this case)
are displayed in real time with a monitor.
{\bf (c)}
Speckle patterns produced by optical scattering from the silicone material. 
The patterns vary depending on the deformation of the silicone material
and temperature. 
{\bf (d)} Proposed deep learning architecture. 
It comprises the common feature extractor and decoder (regression) 
sections. 
}
\end{figure}

{\bf Proof-of-concept experiment.}
We performed an experiment for verifying the proposed sensing approach 
(See Methods for details). 
The sensing targets of the experiment were tactile and thermal sensations.
We chose a commercial transparent silicone elastomer material
as the sensing material [Fig.~\ref{fig_exset}(a)].
The laser light is incident on the silicone material and scattered by impurities or less-visible air bubbles inside the material.
The scattered intensity distribution at $\rr$ on an observation plane $I_{\xx}(\rr)$ 
was measured using a digital camera [Fig.~\ref{fig_exset}(b)]. 
In this experiment, 
a stainless cylindrical indenter was used to apply a normal force 
to the silicone elastomer. 
The position of the indenter was controlled with a positioning stage, 
and the indentation depth and contact location of the indenter were measured. 
The indentation depth is related to the force applied to the silicone material 
(Supplementary Fig.~1); therefore,
we use the indentation depth as a substitute for the applied force.
The measured speckle pattern $I_{\xx}(\rr)$ changes sensitively depending on the indentation depth, location of the indenter, and temperature, as shown in Fig.~\ref{fig_exset}(c).  
The temperature-dependence of the speckle pattern can be attributed to  thermal expansion/contraction or change in the refractive index. 
The features representing the physical stimuli, which are embedded in the speckle patterns,
can be visualized using a nonlinear dimensionality reduction technique \cite{JMLR:v9:vandermaaten08a} (Supplementary Fig.~2). 

{\bf Network architecture.}
Figure~\ref{fig_exset}(d) shows the neural network model to infer $\GG^{-1}(\xx)$ and 
simultaneously estimate $\xx = (x_1,x_2,x_3)$, where $x_1$, $x_2$, and $x_3$ 
correspond to the indentation depth, position of contact along a line in a sensor coordinate, 
and temperature, respectively.
The network model comprises two components: (i) a common feature extractor, which extracts relevant common features from the speckle images and (ii) a decoder (regression model) to transform the extracted feature into physical quantities, $\xx$.
The common feature extractor mainly consists of two convolution neural network (CNN) layers and a single fully connected (FC) layer, whereas 
the decoder consists of branched FC layers for the transformation 
into each physical quantity. %
See Methods for the detailed structure.

\begin{figure}[htbp]
\centering\includegraphics[width=7cm]{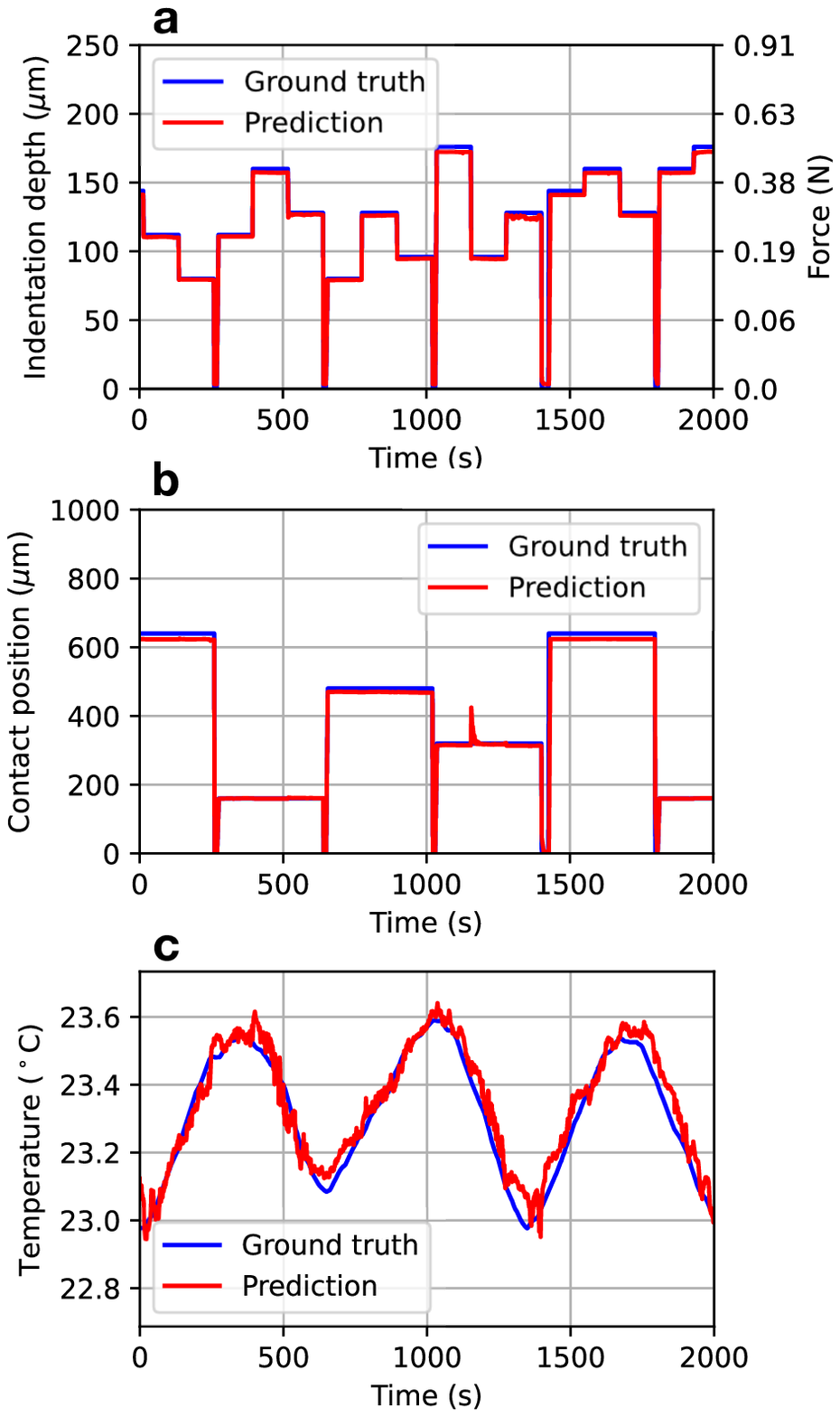}
\caption{\label{fig_time}
{\bf Multimodal sensing demonstration.}
Simultaneous estimations of
{\bf (a)} indentation depth, {\bf (b)} contact position, and {\bf (c)} temperature of the silicone material. 
In {\bf (a)}, the applied force corresponding to the indentation depth 
is also shown. 
In {\bf (b)}, an instantaneous large error at 1156 s is mainly attributed to an unintentional deviation from the set position of the indenter. 
The sensing data points were sampled at time steps of approximately 1.6 s. 
}
\end{figure}

\begin{figure}[htbp]
\centering\includegraphics[width=9cm]{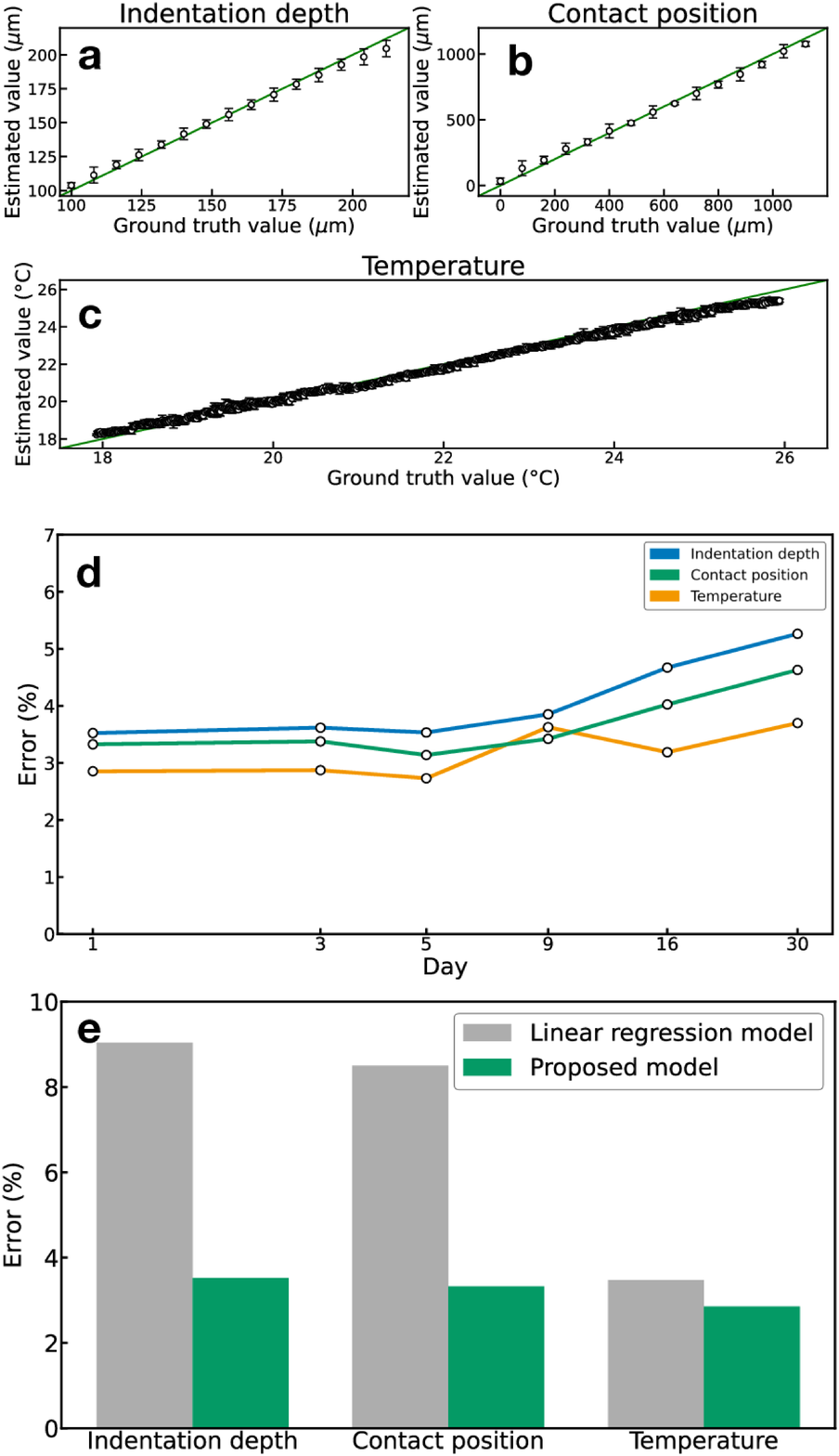}
\caption{\label{fig_result1}
{\bf Sensing performance.}
With our sensing approach, we simultaneously estimated 
the indentation depth, contact position, and temperature of the silicone material. 
The estimated values of 
{\bf (a)} the indentation depth, {\bf (b)} contact position, and {\bf (c)} temperature 
were compared to the ground truth values. 
The error bars represent standard deviations. 
{\bf (d)} Long-term stability of our sensing approach. 
The increase in the estimation errors 
was within 1.7 $\%$ for 30 days after the training. 
{\bf (e)} Model comparison. The estimation errors of the proposed model shown in Fig.~\ref{fig_exset}(d) are compared with a linear regression model, which does not contain hidden layers. 
}
\end{figure}

{\bf Multimodal sensing.}
The network model was trained with $N$ training data samples $\{I_{\xx}^{(n)}(\rr),\xx^{(n)}\}_{n=1}^N$ (see Methods), and used for multimodal sensing of the indentation depth, contact position, and temperature.   
Figure~\ref{fig_time} demonstrates the multimodal sensing,  
where the network model was trained with $N  =4000$ training samples. 
%
%
The simultaneous estimations of $x_1$, $x_2$, and $x_3$ 
can be achieved with a latency of 
only a few hundred milliseconds (Supplementary Fig.~3), even when the depth and location of the indentation 
change in a random manner and the surrounding temperature varies under the effect of an air conditioner. 
Then, the network model was further trained with $N = 15,360$ training samples for better estimation. 
The performance of the simultaneous estimation was evaluated with $18,000$ test cases, 
where the indentation depth $x_1$ and contact position $x_2$ were changed
at intervals of 8 $\mu$m and  80 $\mu$m, respectively, 
and the silicone material was used at room temperature .
The excellent estimation results are summarized in Fig.~\ref{fig_result1}(a)-\ref{fig_result1}(c). 
The estimation errors for the indentation depth, position, and temperature were $\pm$3.95 $\mu$m 
(corresponding to $\pm$25 mN in the range from 0 N to 1.2 N, Supplementary Fig.~1), $\pm$37.25 $\mu$m, and $\pm$0.23 $^\circ$C, 
respectively. 
The relative errors, defined as 
$\langle |x_{i}^{(n)} - \hat{x}_{i}^{(n)}| \rangle/(x_{i,max}-x_{i,min}) \times 100$ ($i =1, 2, 3$), 
were estimated as 3.52 $\%$, 3.33 $\%$, and 2.85 $\%$, respectively, where
$\hat{x}_i^{(n)}$ is the estimated value for $x_i^{(n)}$ of the $n$-th sample, 
and $\langle \cdot \rangle$ denotes the sample mean. 
The errors were close to the precision of positioning of the indenter and temperature controls 
used in this experiment. 

To evaluate the long-term stability of the proposed sensing approach, we recorded the time transition of the estimation errors for 30 days, as shown in Fig.~\ref{fig_result1}(d).
In this experiment, we used the training dataset acquired on the first day and set the network parameters of the model shown in Fig.~\ref{fig_exset}(d); 
then, we measured the estimation errors. 
Although the speckle pattern measurement is generally sensitive to environmental changes, particularly temperature change, our results reveal that the error increases only by 1.7 $\%$ over 30 days and does not significantly change for 5 days, suggesting the robustness of our measurement method.  
This is attributed to the fact that the network architecture is trained for temperature changes. 

To investigate the performance of the proposed deep learning model shown in Fig.~\ref{fig_exset}(d), we compared the model 
with a simple linear regression model, which does not contain hidden layers. 
In the linear regression model, the output vector $\xx$ is produced directly 
from the weighted summation of the input speckle images.
As shown in Fig.~\ref{fig_result1}(e), 
the estimation errors of the proposed model are smaller than those of the regression model, 
suggesting the effectiveness of the CNN-based common feature extraction from the hidden layers in the proposed model. 
Thus, the proposed model is superior to the linear regression model.  

\begin{figure}[htbp]
\centering\includegraphics[width=7cm]{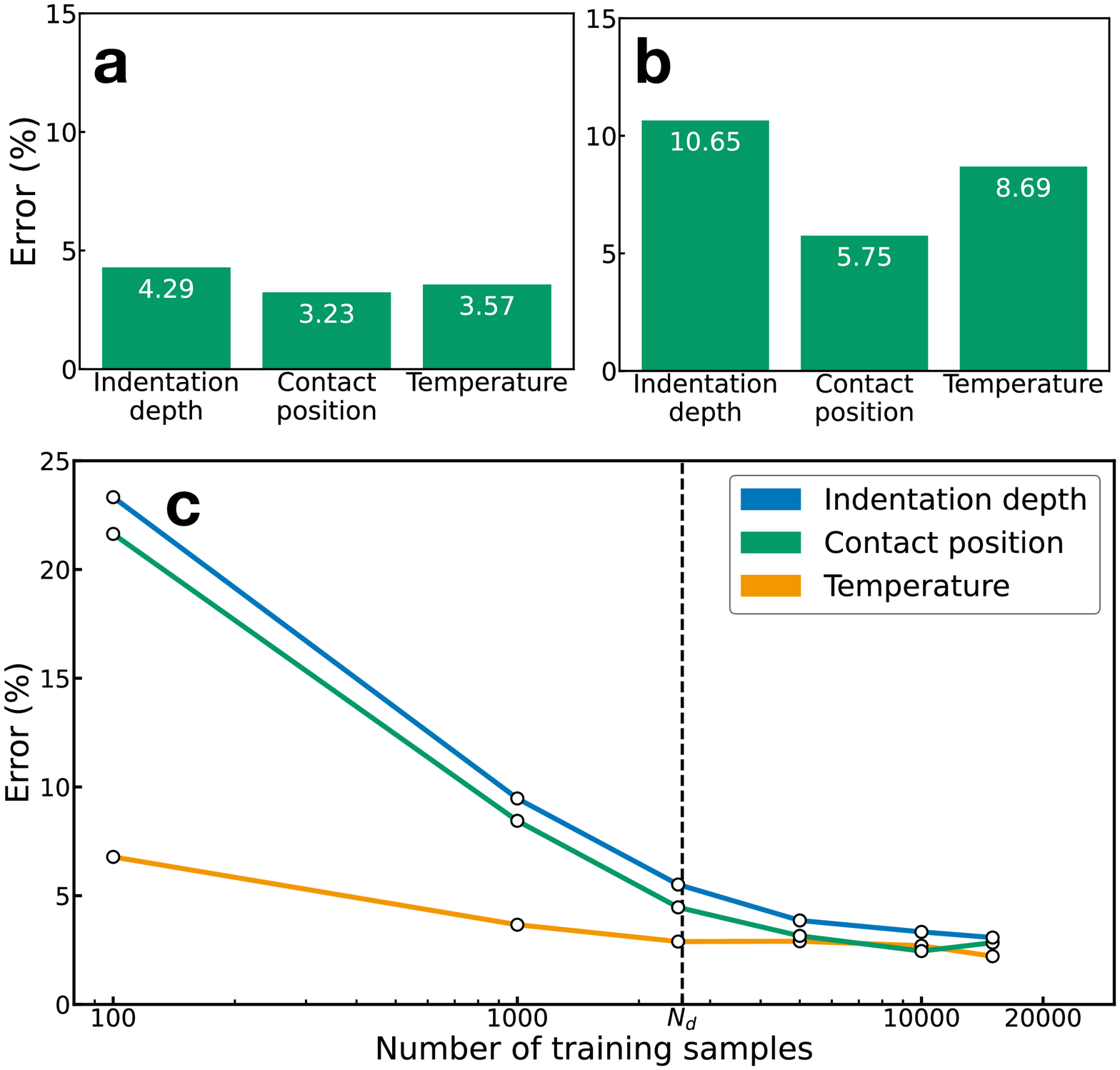}
\caption{\label{fig_result2}
{\bf Effects of training data collection.}
Estimation errors for {\bf (a)} $C(\Delta x_1) \approx 0.61$ and $C(\Delta x_2) \approx 0.66$ 
and {\bf (b)} $C(\Delta x_1) \approx 0.42$ and $C(\Delta x_2) \approx 0.36$.
The estimation errors depend on the sampling intervals $\Delta x_i$ (and the correlation $C(\Delta x_i)$ between the sampled speckle patterns). 
The estimation errors tend to be lower for smaller sampling intervals and higher correlations. 
{\bf (c)} Estimation errors as a function of the number of training samples. 
}
\end{figure}

Training data collection, i.e., the method for collecting the training data samples, 
is crucial for realizing excellent estimations using the proposed sensing approach. 
Considering that the regression (interpolative estimation) is based on the training using correlated samples, the sampling interval, $\Delta x_i$, for the training data of $x_i$ ($i=1, 2, 3$) 
should be tuned such that the speckle patterns are correlated between the training samples.
If the sampling interval is too large for the training samples to be correlated,
the interpolative estimation between the sampling intervals will be generally difficult. 
To investigate the effect of the sampling interval on the estimation performance, 
we changed the sampling intervals, $\Delta x_1$ and $\Delta x_2$, for the indentation depth and contact position
and characterized the speckle correlation as 
$C(\Delta x_i) = \langle(I_{x_i}-\bar{I}_{x_i})(I_{x_i+\Delta x_i}-\bar{I}_{x_i+\Delta x_i})\rangle/(\sigma_{x_i}\sigma_{x_i+\Delta x_i})$ ($i=1, 2$), 
where $\bar{I}_x$ and $\sigma_x$ are the mean and standard deviation of the speckle intensity pattern, respectively. 
As shown in Fig.~\ref{fig_result2}, the estimation errors depend on the speckle correlation $C(\Delta x_i)$. 
For $\Delta x_1$ = 12 $\mu$m and $\Delta x_2$ = 120 $\mu$m, the speckle correlations were $C(\Delta x_1) \approx 0.61$ and $C(\Delta x_2)$ $\approx 0.66$, 
respectively.
In this case, the estimation errors were 4.29 $\%$, 3.23 $\%$, and 3.57 $\%$ [Fig.~\ref{fig_result2}(a)].
When the sampling intervals are very large ($\Delta x_1$ = 52$\mu$m and $\Delta x_2$ = 520 $\mu$m), $C(\Delta x_1)$ and $C(\Delta x_2)$ decrease to 0.42 and 0.36, respectively, and 
the mean estimation error increases to approximately 10.65 $\%$, 5.75 $\%$, and 8.69 $\%$ 
[Fig.~\ref{fig_result2}(b)], which suggests that 
the speckle correlation between the samples affects the generalization capability of the proposed model.

It is also important to consider the appropriate training data size for good estimation performance with low training cost.
Figure~\ref{fig_result2}(c) shows the estimation errors as a function of the number of training samples, $N$. 
As seen in this figure, the estimation errors sufficiently decrease when $N > N_d = 2560$, indicated by the vertical dotted line, where 
$N_d$ is the number of training samples that cover all states of the soft material in response to indentation and temperature changes in this experiment (see Methods). 
The oversampling for $N > N_d$ affects the further lowering of the errors, suggesting the compensation  of the errors caused by laser noise and temperature fluctuation.

\begin{figure}[htbp]
\centering\includegraphics[width=12cm]{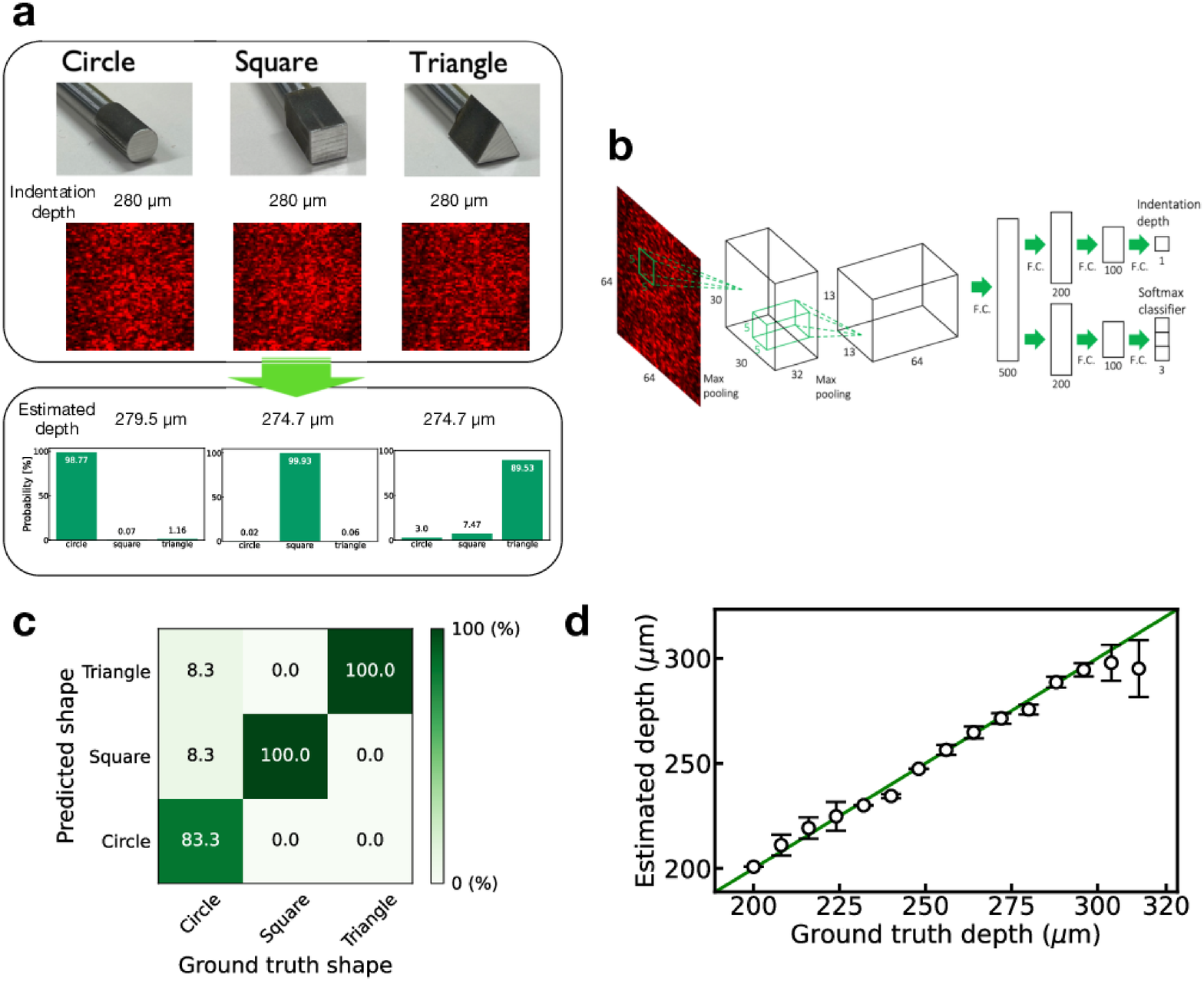}
\caption{\label{fig_class}
{\bf Simultaneous realization of sensing and perception.} 
{ \bf (a)} Three indenters with circular, square, and triangular cross sections 
were used in the experiment. 
The speckle patterns change depending on the shapes of the indenters and contact force on the soft material. 
Examples of the speckle patterns 
for each indenter with the indentation depth of 280 $\mu$m are shown in { \bf (a)}. 
The indenter shape and depth can be identified in the network model shown in {\bf (b)}, 
where the common feature extractor is the same as that in the model shown in Fig.~\ref{fig_exset}(d). 
The network can be trained to reduce both the mean squared error and cross entropy
with 405 training samples. 
The estimation results for 45 test samples are shown in {\bf (c)} and {\bf (d)}. 
{\bf (c)} Confusion matrix of shape identification.
{\bf (d)} Estimation of the indentation depth.
}
\end{figure}

{\bf Sensing and perception.}
As mentioned above, the speckle patterns are highly sensitive to various external stimuli on the soft material, i.e., they contain the information of various external stimuli. 
This suggests that the information of multiple external stimuli, which is difficult to detect using a single-mode sensor (e.g., pressure sensor and temperature sensor), can be extracted from the speckle patterns.
In addition, this results in the advantage that 
it is not necessary to rebuild the sensing device when changing the target stimuli to sense different physical quantities; instead, it is sufficient to modify the training and post-processing of the sensing in our sensing approach.  
As a demonstration, we selected various shapes of the indenters used to deform the soft material as the sensing targets and verified the identification of 
the shapes of the indenters along with the sensing of the indentation depth (corresponding to the contact force).
In the experiment, we used three types of indenters with 
circular, square, and triangular cross-sections, respectively [Fig.~\ref{fig_class}(a)]; the areas of the cross-sections were equal to each other. 
Although shape identification generally requires spatial information of the deformation, 
which cannot be detected using a single-point measurement, 
our sensing approach can optically grasp the spatial information with an ultrahigh spatial 
resolution of few tens of micrometers (Fig.~\ref{fig_result1}). 
The shape identification can be easily achieved using a shape classifier 
in the decoder section of the network architecture, 
as shown in Fig.~\ref{fig_class}(b).
The network model can be trained such that the mean squared error for the regression 
of the indentation depth and cross entropy for the classification of the indenter shape 
are both minimized. 
Figures~\ref{fig_class}(c) and \ref{fig_class}(d) show the simultaneous estimation results of the indenter shapes and indentation depths (also, the indentation force). The experiment used 450 samples
(90$\%$  and 10$\%$ of the total samples were used for training and testing, respectively). 
The identification error was approximately 94$\%$, and the mean 
estimation error of the indentation depth was approximately 3.8$\%$ 
(corresponding to a precision of 4.2 $\mu$m).
The network model is highly scalable for the number of estimated physical quantities; thus, using a larger number of training samples can enable simultaneous estimation of additional physical quantities with lower errors.  

\begin{figure}[htbp]
\centering\includegraphics[width=10cm]{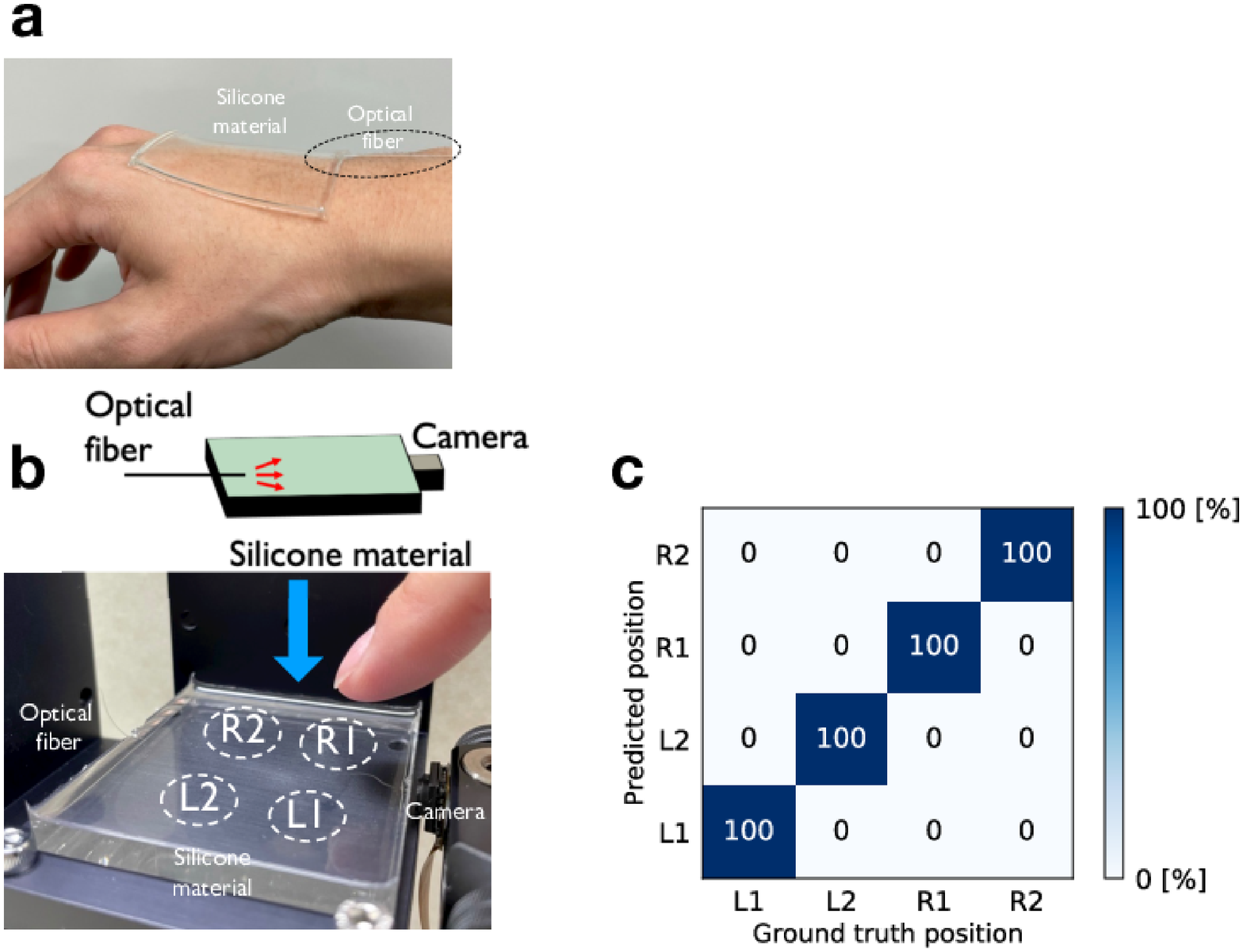}
\caption{\label{fig_kitagawa}
{\bf Human--machine interface.} The interface device consists of a thin transparent soft material, optical fiber, and miniature camera. 
{ \bf (a)} A thin silicone material can be fitted on the human hand. 
{\bf (b)} Interface device. %
The device can detect four positions, indicated by $L1$, $R1$, $L2$, and $R2 $, where the silicone material is pressed by a human finger. 
{\bf (c)} Classification accuracy. The contact position can be identified with an accuracy of 100 $\%$ after training.
}
\end{figure}

{\bf Toward human--machine interface.}  
For application in a human--machine interface, 
the proposed optical sensing unit can be easily incorporated 
with an optical fiber to deliver the laser light to the soft material
and a miniature camera for detection.
A thin skin-like silicone material can be fitted with the human body to allow 
physical sensing and controlling [Fig.~\ref{fig_kitagawa}(a)]. 
More importantly, the proposed optical sensing approach allows us to  
control the sensitivity for detecting the external stimuli via speckle patterns.
We attempted to change the sensing precision from the micrometer scale, demonstrated above, to the millimeter scale, which is more suitable for detecting touching motions.
Such a sensitivity reduction can be achieved by suppressing high-order multiple 
scattering and reducing the camera sensitivity, i.e., mainly detecting 
single scattering or low-order multiple scattering. 
Figure~\ref{fig_kitagawa}(b) shows our proposed soft interface device, which 
consists of a transparent silicone material, optical fiber, and miniature digital camera. 
The laser power and exposure time of the camera 
were tuned such that only a single scattering or low-order multiple signals, which are less sensitive to the deformation, could be detected. 
The silicone material was pressed by a human finger 
at four positions, indicated by $L1$, $L2$, $R1$, and $R2$, 
and the speckle patterns were measured using a miniature camera. 
The contact positions were successfully identified 
after the training using 320 samples [Fig.~\ref{fig_kitagawa}(c)].
The excellent result is due to the simple task of classifying the stimuli into a few classes instead of considering the multimodal regression problem.

\section*{Discussion}
In this work, we demonstrated an optical multimodal sensing approach, which enables 
highly sensitive simultaneous sensing of physical contact and temperature changes,
with additional advantages of low electromagnetic interference, 
noninvasive nature, high stability, and low-cost design.
The high sensitivity originates from the optical interference techniques, which are based on the responses of the speckle patterns to physical stimuli.
The sensing mechanism is entirely different from previous optical sensing approaches, which used the changes in the light intensity and wavelength.
We emphasize that the sensing approach can be applied to various optical materials because the only prerequisite is the occurrence of a speckle phenomenon in the sensing material. 
Owing to the high sensitivity realized with the optical interference techniques, 
the proposed approach can enable the sensing of physical contacts with high spatial resolution of few tens of micrometers. 
Thus, it outperforms previous electrical and optical approaches.
Higher spatial resolution and greater precision of measurement of the indentation can be achieved by using a larger number of training samples.
An opaque material, which induces higher-order multiple scattering, 
can facilitate speckle-based sensing with greater sensitivity.  

Regarding the implementation of the multimodal sensing principle,
we emphasize that the proposed approach is superior to previous multimodal sensing approaches, as it avoids complex integration of multiple sensing elements with different configurations or designs of sensor networks (sensor matrices) into a single sensing device with numerous wire connections.
This unique property would be useful in seamlessly integrating the skin-like sensors with soft actuators for realizing soft robotics.
%

Despite the advantages of the proposed approach, there is scope for further improvement. 
The main drawback of the proposed approach is the requirement of a large-scale model with a large number of training samples for multimodal sensing, which may make real-time operation difficult when multiple parameters need to be estimated.  
One way of overcoming this drawback is to prepare an ensemble of models that are separately trained for different datasets and to distill the knowledge to be transferred 
from the trained model to a small model. 
This may be possible by adopting a transfer learning technique or 
knowledge distillation \cite{Hinton2015DistillingTK}.
Knowledge distillation also enables the small model to be 
trained on much less data \cite{Hinton2015DistillingTK}.
%

Because of the high information capacity of the speckle patterns, the proposed approach will enable higher-level perception and higher multimodality, including the multimodal sensing of shear force, distortion, and three-dimensional shapes of a contact object, 
which are difficult to sense using conventional sensors but are important when considering the grasp of soft and slippery objects such as robot hands.
Our model-free (data-driven) sensing strategy allows us to freely choose the sensing modes according to our purpose. 
An interesting topic for future work will be to realize soft sensing with nanometer-scale spatial resolution. Another interesting future work will be to create sensory signals for fuzzy sensations such as touch, warmth, or pain for use in prosthetic sensory skin by integrating and processing multimodal sensory signals.

\section*{Methods}

{\bf Experimental setup.} 
A schematic of the proof-of-concept experiment is shown in Supplementary Fig.~4. 
We used a commercial transparent soft silicone material (Verde Co., Ltd., Superclear silicone). 
The physical dimensions of the silicone material were 58 mm$\times$52 mm$\times$5 mm. 
A He-Ne laser (MELLES GRIOT, 05-LHP-991, Wavelength 632.8 nm) 
was used as the light source. 
The scattered intensity distribution was detected with a digital camera 
(Thorlabs, DCC1240C) with exposure time of 5 ms.  
The silicone material was deformed using a stainless cylindrical indenter with 
contact area diameter of 3 mm.  
The depth $x_1$ and location $x_2$ of the contact with the indenter were controlled using a two-axis stepping motor controller with precision of $\pm 0.5 \mu$m.
We determined the origin of the indentation depth by moving the indenter until the contact could be observed.
The contact position $x_2$ was moved along a vertical line. 
The temperature $x_3$ of the silicone material was controlled using a Peltier device with a thermistor embedded inside the silicone material for temperature monitoring [Supplementary Fig.~4]. (The precision of temperature control was estimated as $\pm$0.2$^\circ$C.)

The sensing unit shown in Fig.~\ref{fig_kitagawa}(b) consists of a  
transparent silicone material, polarization-maintaining optical fiber (Thorlabs, PM630-HP), and miniature digital camera with angle view of 62$\times$47 degrees  (ArduCam, OV5647 Spy Camera Module for Raspberry Pi). 
The total input power of the laser light was less than 1 mW, 
and the exposure time of the camera was set as 1/8 s. 

{\bf Preprocessing and network model.}
For preprocessing of the measured speckle images to be used as input images 
to the network model shown in Fig.~\ref{fig_exset}(d), 
the images were downsampled to 30 $\%$ and trimmed to 
64$\times$64 pixel images. 
The kernel size of the downsampling 
was set to be close to the mean speckle size in order to reduce the sensitivity of the speckle patterns to environmental fluctuations such as vibration and air fluctuation.
The resizing effects on the estimation errors are shown 
in Supplementary Fig.~5. 
Regarding the physical parameter vector $\xx = (x_1,x_2,x_3)$, 
each feature $x_1$, $x_2$, and $x_3$ were scaled to remove the difficulty due to different physical dimensions such that 
they were in the range between -1 and 1 in the training dataset, 
i.e., $x_i \rightarrow x_{s,i} = 2(x_i-x_{i,min})/(x_{i,max}-x_{i,min}) - 1$, $(i = \{1,2,3\})$,
where $x_{i,max}$ and $x_{i,min}$ are the maximum and minimum values of $x_i$ 
in the training dataset, respectively. 
Let $\xx_s = (x_{s,1},x_{s,2},x_{s,3})$ be the scaled vector, and 
by using a training dataset of $N$ cases, $\{I^{(n)}_{\xx}(\rr),\xx_s^{(n)}\}_{n=1}^N$, 
the network parameters are trained so that the following mean square error (MSE) is minimized:
MSE $= 1/N\sum_{n=1}^N(\xx_s^{(n)}-\hat{\xx}_s^{(n)})^2$, 
where $\hat{\xx}_s^{(n)}$ is the output vector for the $n$th sample.
The inferred parameter vector $\hat{\xx}^{(n)}$ can be obtained from the inverse transformation of the min-max scaling, $\hat{\xx}_s^{(n)} \rightarrow \hat{\xx}^{(n)}$. 
The mini-batch size for the gradient-based optimization was set as 50, and the Adam optimizer was used.

In the deep learning model shown in Fig.~\ref{fig_exset}(d), 
the first CNN layer uses 32 kernels of size $5\times 5$ 
and the ReLU activation function, followed by batch normalization 
and max pooling of size $2 \times 2$ and stride 2.  
The second CNN layer uses 64 kernels of size $5 \times 5$ and
the ReLU activation function, followed by batch normalization and 
max pooling of size $2 \times 2$ and stride 2.  
The features are shared in three branched networks, which consist of three FC layers. 
FC matrices of size $500 \times 200$, $200 \times 100$, and $100\times 1$ are used in 
the first, second, and third FC layers, respectively.

In the deep learning model shown in Fig.~\ref{fig_class}(b), the common feature extractor 
is the same as in the model shown in Fig.~\ref{fig_exset}(d).
In the decoder section, a classifier was added for shape identification. 
The network was trained such that both the MSE for the regression and cross entropy for the classification were minimized. 

{\bf Data collection protocol.}
To collect the data to train the network model shown in Fig.~\ref{fig_exset}(d), we changed the indentation depth from $x_1 = $100 $\mu$m to 212 $\mu$m with a sampling interval of $\Delta x_1 =$ 16 $\mu$m, the contact position from $x_2 = $ 0 $\mu$m to 1120 $\mu$m with an interval of $\Delta x_2 = $ 160 $\mu$m, temperature from $x_3 = $17.9$^\circ$C to 26.0$^\circ$C with an interval of 0.2 $^\circ$C, where the origin of $x_1$ (pressing depth of the indenter) was set as the surface of the silicone material. 
We simultaneously recorded the speckle pattern $I_{\xx}(\rr)$ for each stimulus. 
The number of training samples obtained in this process, $N_d$, 
was $\sim$2560.  
We repeated this sampling process and set $N/N_d \approx 6$ to reduce the fluctuation in the measured speckle patterns and improve the robustness of the network to noise, 
where $N(\approx 6N_d) = 15,360$ is the total number of training samples used for the proof-of-concept experiment (Fig.~\ref{fig_result1}).

\section*{Data availability}
The data that support the findings of this study are available 
from the corresponding author upon reasonable request.

\bibliographystyle{apsrev}
\bibliography{/Users/sunada/My_Refs}

\section*{Acknowledgments}
This work was partly supported by JSPS KAKENHI (Grant No.~20H04255) 
and JST PRESTO (Grant No.~JPMJPR19M4).
The authors thank J. Hanawa for the numerical computation 
of dimensionality reduction and Y. Miyajima for helpful support on the compression test of silicone elastomer. 

\section*{Author contributions}
Sa.S. conceived the idea and directed the project. 
Sh.S. designed the experimental setup and 
performed the experiments and numerical analysis. 
K.S. also performed the numerical analysis. 
K.K. performed the experiment and numerical analysis shown in Fig.~\ref{fig_kitagawa}. 
Sa.S., Sh.S., and T.N. discussed and wrote the manuscript. 
All authors contributed to the preparation of the manuscript. 

\section*{Competing interests}
The authors declare no competing interests. 

%

\setcounter{figure}{0}
\section*{Supplementary information}

\section*{Mechanical properties of the silicone material}
A compression test for the silicone material used in this study was performed at a constant speed of 0.29 mm/min at room temperature. 
The compressive elastic modulus was calculated from the strain and nominal stress by focusing on the linear response regime. 
The elastic modulus of compression was in the range of 3--10 MPa. 

\begin{figure}[htbp]
\centering\includegraphics[width=8cm]{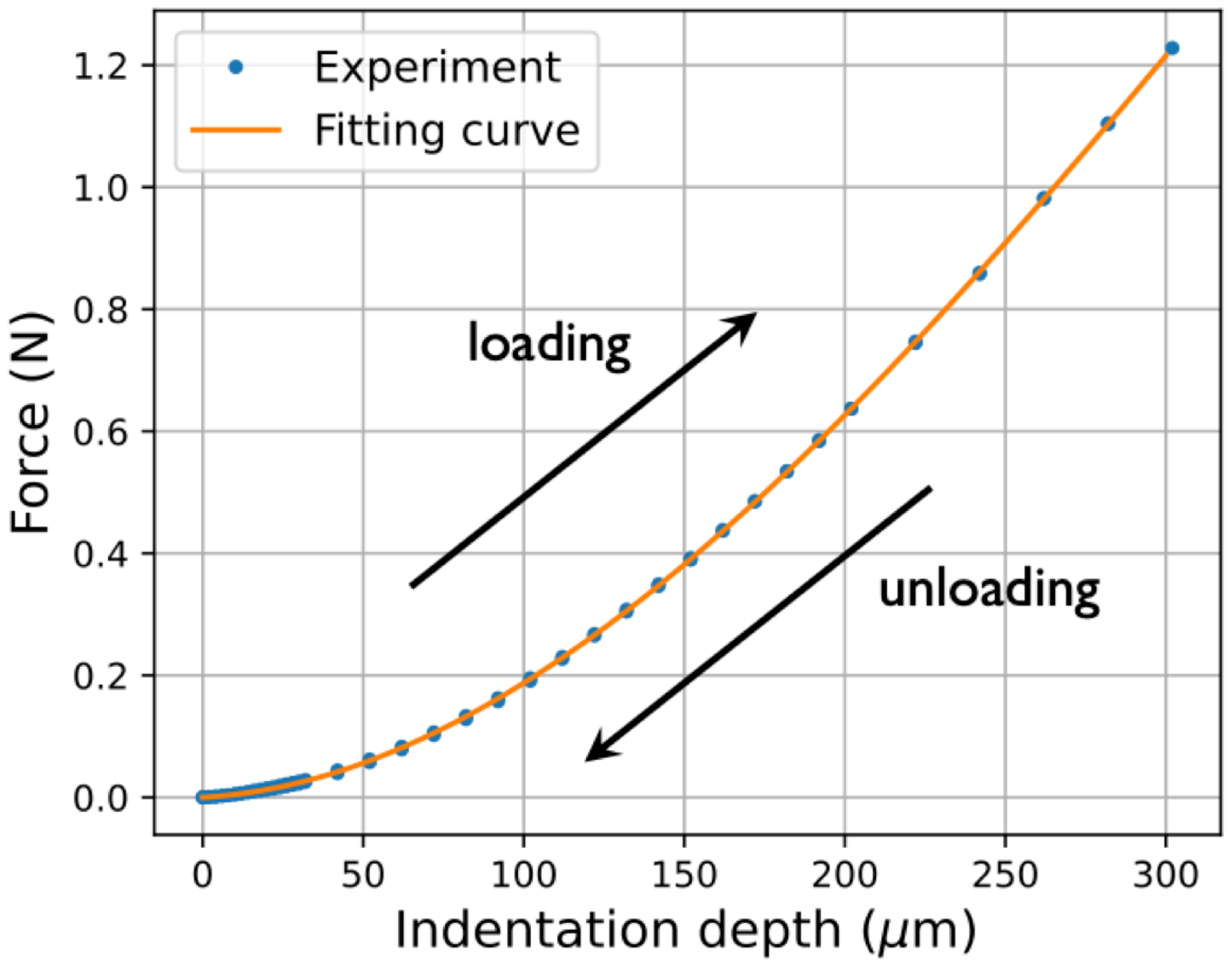}
\caption{\label{fig_force}
Applied force as a function of indentation depth for the silicone material in the loading and unloading regimes. 
}
\end{figure}

We also performed an indentation experiment for the silicone material in both loading and unloading regimes.
In this test, a stainless cylindrical indenter with 
the contact area diameter of 3 mm was used. 
The result is shown in Fig.~\ref{fig_force}, where the force applied to the silicone material is displayed as a function of the indentation depth. 
Note that the hysteresis between the loading and unloading regimes in the experimental result was not significant. 
We fitted the experimental data with a fourth-order polynomial curve. 
The root mean square error was 0.0016. 
We estimated the applied forces corresponding to the measured indentation depth (shown in the main text) by using the fitting curve. 

\newpage
\section*{ Two-dimensional projection of speckle images}

\begin{figure}[htbp]
\centering\includegraphics[width=10cm]{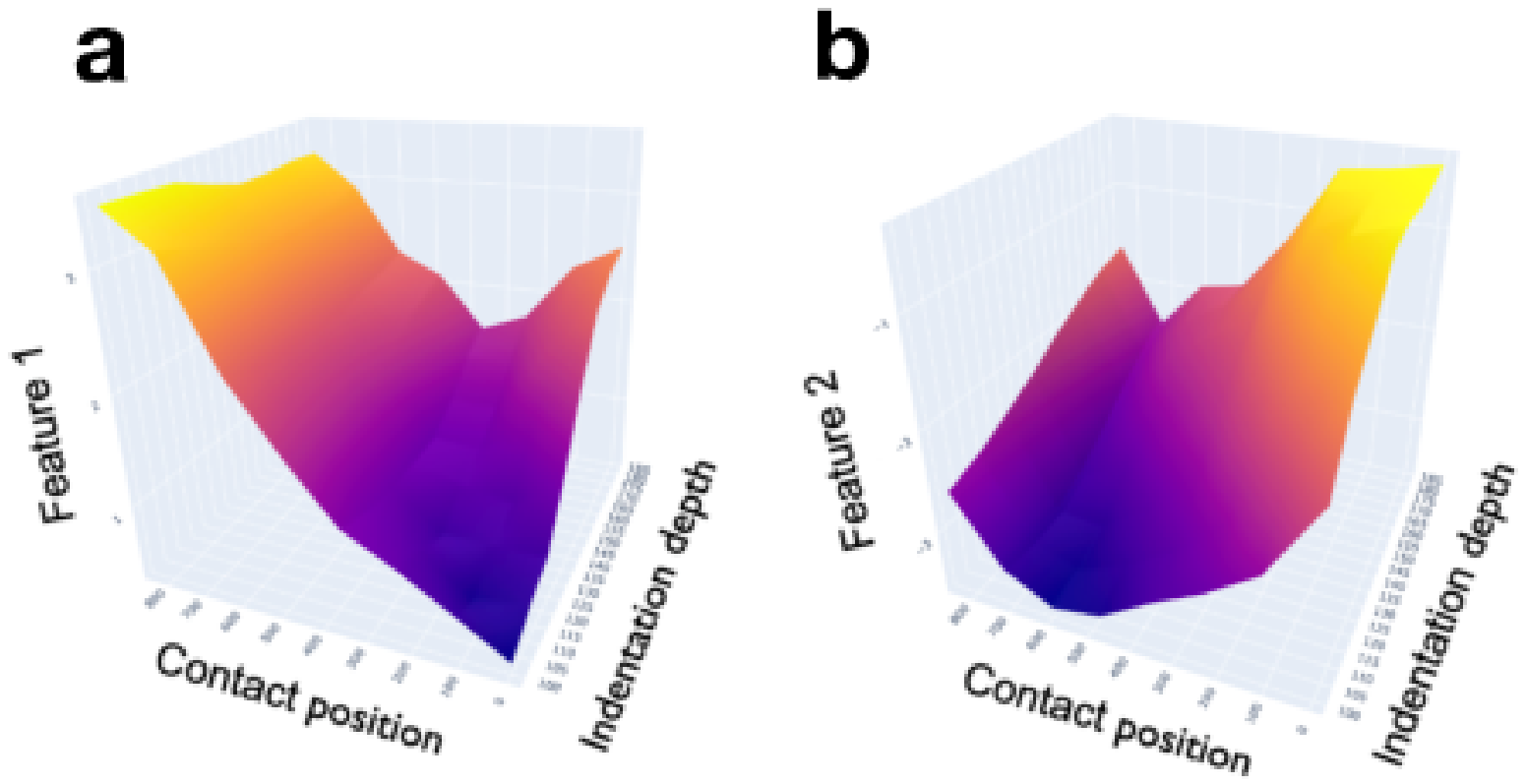}
\caption{\label{fig_SI_tsne}
{\bf Two-dimensional projection of the speckle image dataset.} 
The speckle images were projected onto a two-dimensional feature space for visualization by a nonlinear dimensionality reduction technique (t-distributed stochastic neighbor embedding, t-SNE).
The perplexity was set to 50. 
In {\bf(a)} and {\bf(b)}, two features 1 and 2 embedded in the speckle patterns are shown as functions of the pressed depth and position.  
The features continuously change depending on the physical stimuli, which 
suggests that they can be extracted from the speckle patterns using a decoder.
}
\end{figure}

\section*{Latency and computation time}
\begin{figure}[htbp]
\centering\includegraphics[width=8cm]{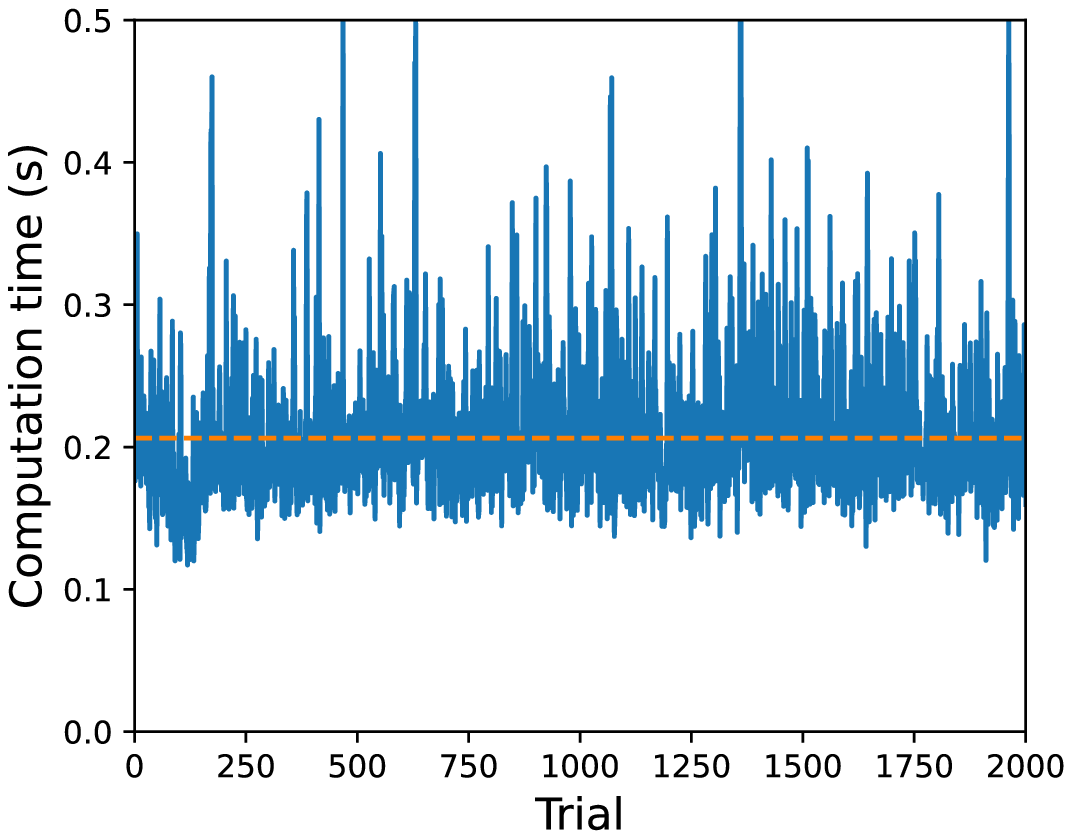}
\caption{\label{fig_SI_comptime}
Computation time for 2000 trials. 
The dashed line denotes the mean computation time of 206 ms. 
}
\end{figure}

The simultaneous estimations of three parameters can be achieved with a latency (time delay) of few hundred milliseconds [Fig.~2 in the main text].
The latency is mainly due to the computation time between the preprocessing of the captured speckle images and output of the proposed model [Fig.~2(d)]. 
We measured the total computation time, including the preprocessing and computation of the proposed model [Fig.~\ref{fig_SI_comptime}]. 
The specifications of the computer used in the experiment are summarized in Table~\ref{tb:pc}.
The mean computation time for 2000 trials was approximately 206 $\pm$ 50 ms.

\begin{table}[h]
 \caption{Specifications of the computer used in the experiment.}
\centering
\begin{tabular}{|c|c|}
\hline
Item & Value \\
\hline
OS & Windows 10 \\
CPU & Core(TM) i7-10700 CPU @ 2.90 GHz  \\
RAM & 16.0 GB \\
GPU & Geforce GTX 1650 SUPER \\
\hline
\end{tabular}
\label{tb:pc}
\end{table}
%

\newpage
\section*{Experimental setup}
\begin{figure}[htbp]
\centering\includegraphics[width=7cm]{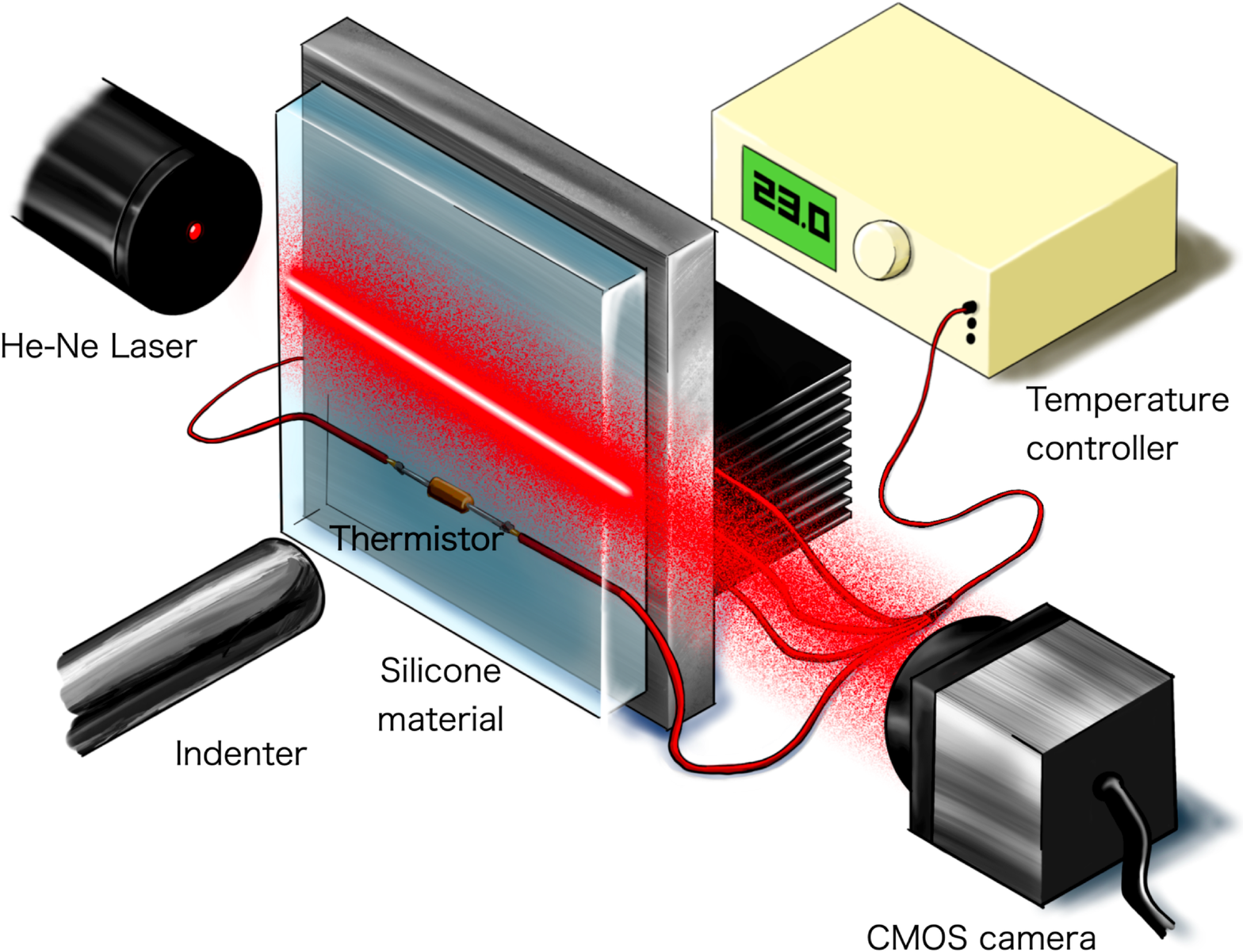}
\caption{\label{fig_SI_expset}
Schematic of the experimental setup for the results shown in Figs.~2-4 in the main text.
}
\end{figure}

\section*{Resizing effect}
\begin{figure}[htbp]
\centering\includegraphics[width=6cm]{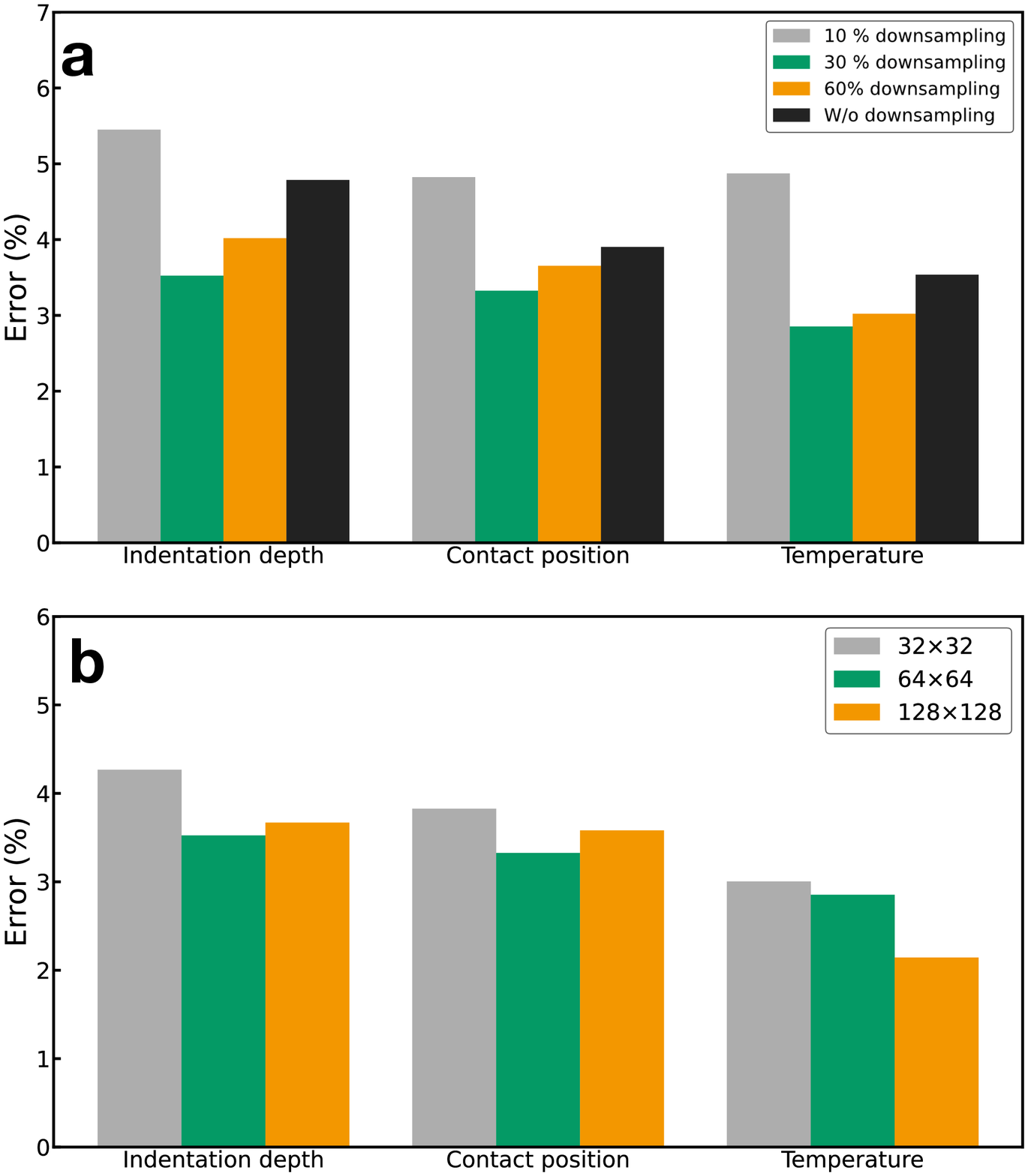}
\caption{\label{fig_SI_samp}
{\bf Resizing effects on estimation error.}
The speckle images were downsampled and trimmed for the pre-processing.  
{\bf (a)} Effect of downsampling. 
The image size was fixed as $64\times 64$-pixel size. 
The errors are minimized when the original images were downsampled to 30$\%$.
The large downsampling of 10$\%$ degrades the estimation performance. 
{\bf (b)} Effect of the image size on the estimation error. 
As the image size is reduced, the computation cost is reduced for the estimation but the performance is degraded.
In this study, we used the speckle images of $64\times 64$ pixel size. 
}
\end{figure}

\end{document}